\documentclass[11pt]{article}

\usepackage[preprint]{acl}

\usepackage{times}
\usepackage{latexsym}
\usepackage{booktabs} 
\usepackage{float} 
\usepackage[T1]{fontenc}

\usepackage[utf8]{inputenc}

\usepackage{microtype}

\usepackage{inconsolata}

\usepackage{graphicx}

\usepackage{amsmath}
\usepackage{amssymb}
\usepackage{multirow} 

\DeclareMathOperator*{\argmax}{arg\,max}

\title{Sparsity-Aware Evolution for Model Merging}

\author{
Huan Zhang\textsuperscript{1,2}\thanks{Equal contribution.},
Yanjian Zhang\textsuperscript{3,4}\footnotemark[1],
Nadi Tomeh\textsuperscript{3},
Guillaume Wisniewski\textsuperscript{4},
Bang Liu\textsuperscript{1,2,5}\thanks{Corresponding author.}
\\
\textsuperscript{1} DIRO \& Institut Courtois, Université de Montréal\\
\textsuperscript{2} Mila -- Quebec AI Institute\\
\textsuperscript{3} Université Sorbonne Paris Nord, LIPN, CNRS\\
\textsuperscript{4} Université Paris Cité, LLF, CNRS\\
\textsuperscript{5} Canada CIFAR AI Chair\\
\{huan.zhang, bang.liu\}@umontreal.ca \\
\{yanjian.zhang, nadi.tomeh\}@lipn.univ-paris13.fr \\
guillaume.wisniewski@u-paris.fr
}

\begin{document}
\maketitle
\begin{abstract}

We propose a sparsity-aware evolutionary (SAE) framework for model merging that involves iterative pruning-merging cycles to act as a novel mutation operator.
We incorporate the sparsity constraints into the score function, which steers the evolutionary process to favor more sparse models, in addition to other conventional performance scores. 
Interestingly, the by-product of \textit{competition} for sparsity introduces an extra local \textit{attraction} and interplay into the evolutionary process: if one competitor has more zero elements, the other competitor's non-zero elements will occupy those positions, even though the less sparse competitor loses to the more sparse competitor in other positions. 
The proposed pipeline is evaluated on a variety of large-scale LLM benchmarks. 
Experiments demonstrate that our approach can improve model merging reliability across multiple benchmarks, and is easy to incorporate due to its simplicity and being orthogonal to most existing approaches. 
The code has been uploaded into the OpenReview system. 

\end{abstract}

\section{Introduction}

Model merging~\citep{yang2024model,ruan2025taskspecific}, also known as model fusion~\citep{li2023deep}, has emerged as an efficient empowerment technique that directly combines the parameters of multiple separately trained models with different capabilities into a single ``universal model'', without requiring access to the original training data or incurring expensive computation.
This approach is possible because deep neural networks share similar low-dimensional parametric subspaces~\citep{kaushik2025universal}. 
Within these universal subspaces, models can be merged to not only aggregate distinct strengths but also to synthesize genuinely new compositional skills, essentially allowing the resulting model to solve complex problems by chaining the atomic skills of its parents~\citep{yuan2025fx}. 

\begin{figure}[t!]
  \centering
  \includegraphics[width=\linewidth]{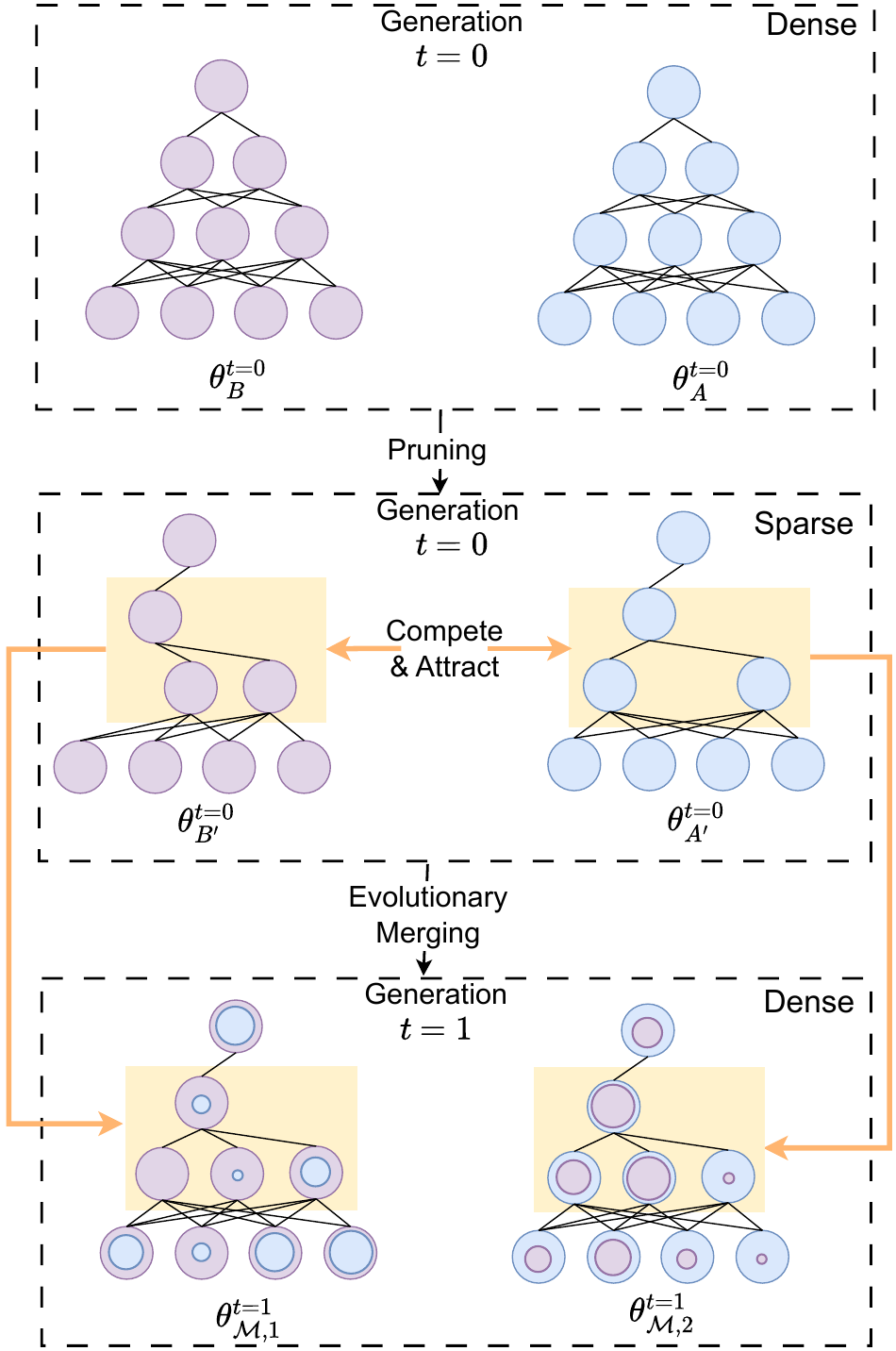}
  \caption{$\theta_A$ and $\theta_B$ are pretrained LLMs that are to be merged into $\theta_{\mathcal{M}}$. 
  Different sizes of circles represent the mixing ratios belonging to different parents. 
  We maintain a large archive of models after generation $t=0$ to promote diversity based on local and global competition mechanisms.
  Note that for the generation $t=1$, the upper-right neuron does not exist, since the parents' corresponding neurons have been pruned in the generation $t=0$. }
  \vspace{-6mm}
  \label{fig:merge}
\end{figure}

Among the diverse strategies for model fusion~\citep{li2023deep}, evolutionary merging approaches have shown particular promise by automating the search for optimal merging configurations in a data-driven manner.
Unlike static averaging methods~\citep{ilharco2023} that rely on fixed heuristics, evolutionary algorithms~\citep{Abrantes2025,Zhang2025}, dynamically explore the vast parameter space of neural models to discover non-intuitive combinations that maximize model performance. This flexibility allows them to adaptively balance the contributions of different parent models, making them highly effective at navigating the complex trade-offs inherent in model fusion without requiring extensive retraining.

\vspace{-2mm}

In this work, we introduce sparsity specifically to enhance these evolutionary merging frameworks, positioning it as a critical regulatory mechanism rather than just a compression tool~\citep{zhu2017prune}, as shown in Figure~\ref{fig:merge}. 
By incorporating sparsity constraints directly into the fitness function of the evolutionary algorithm, we induce a dual dynamic of competition and attraction: the drive for sparsity forces parameters to compete for limited ``survival slots'', essentially pruning redundant or conflicting weights (as detailed in Section \ref{sparsity}), while simultaneously creating a natural attraction where the zeroed-out regions of one model are seamlessly occupied by the active parameters of another (as detailed in Section \ref{attraction}). This synergy steers the evolutionary search toward cleaner, more modular solutions that are less prone to interference.

Standard weight merging often suffers from destructive interference, a phenomenon where conflicting parameter updates across tasks cancel out specialized capabilities, leading to sub-optimal performance \citep{Farajtabar2020, yadav2023ties}. 
To mitigate this within an evolutionary framework, we propose leveraging sparsity \citep{Blalock2020} not merely as a static regularizer against overfitting \citep{Srivastava2014}, but as an active \emph{selection pressure} for conflict resolution. 
By pruning conflicting regions during evolution, we force the algorithm to resolve parameter clashes dynamically. Crucially, this is followed by a re-dense phase, where the space cleared by sparsity is strategically repopulated with complementary features from other models, allowing distinct functional experts to coexist without overwriting each other.

This cycle of sparsification and re-densification transforms the model from a monolithic weight block into a modular landscape of specialized subspaces. By isolating atomic subnetworks—analogous to functional circuits in mechanistic interpretability \citep{olah2020zoom, yuan2025fx}, we prevent noise from irrelevant parameters from disrupting the delicate reasoning chains required for complex tasks. 
This isolation is particularly effective for evolutionary merging, as it provides the search algorithm with cleaner building blocks, ensuring that the merged model can effectively compose skills from different parent models as modular units rather than entangled weights \citep{Elmoznino}.

Furthermore, sparsity serves as a navigational constraint within the shared low-dimensional parametric subspace where effective solutions typically reside \citep{kaushik2025universal,Zhang2025}. 
While dense optimization in the full parameter space is prone to drifting into redundant or harmful regions, our sparsity-guided evolutionary search restricts the process to these essential manifolds. 
By alternating between pruning to maintain structural integrity and re-densing to maximize capacity, we ensure the merged model evolves within the most generalizable regions of the parameter space. 


\begin{figure}[t!]
  \centering
  \includegraphics[width=0.6\linewidth]{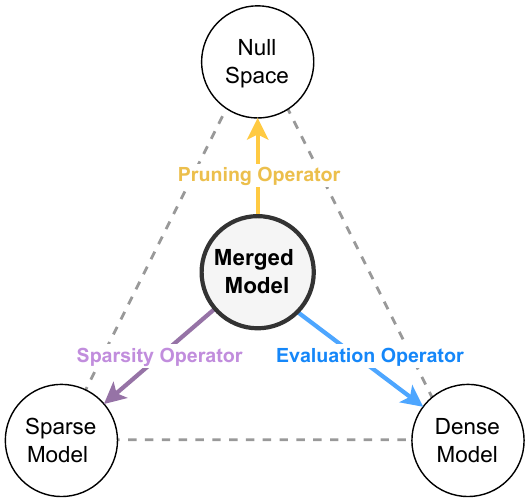}
  \caption{Evolutionary forces in sparsity-aware model merging.
  Evaluation and sparsity jointly act as a natural selection mechanism over offspring models,
  while pruning introduces directed exploration toward increasingly empty parameter regions.
  The merged model evolves within the space spanned by dense model, sparse model, and null space.}
  \label{fig:force}
  \vspace{-3mm}
\end{figure}

Our framework can also be understood from a dynamic perspective.
In particular, the joint use of evaluation and sparsity-aware objectives functions as a form of natural selection over offspring models, favoring solutions that achieve strong performance with fewer active parameters.
Pruning operations complement this selection pressure by enabling directed exploration toward increasingly empty regions of the parameter space.
As illustrated in Figure~\ref{fig:force}, the interaction of these forces drives the merged model to continuously move and search within the space spanned by dense model, sparse model and null space, rather than collapsing to any single extreme.

A na\"ive way would be to constrain the representation capacity of the model merging space, but this requires manually incorporating appropriate priors about the task at hand. 
Inspired by the effectiveness and simplicity of sparsity mechanisms (e.g., dropout and pruning) to reduce overfitting, we design a sparsity-inducing evolutionary approach to regulate model merging space in a data-driven manner.





Our contributions are as follows:
\begin{itemize}
    \item We propose a sparsity-aware evolutionary (SAE) framework that seamlessly integrates sparsity as a direct regulatory signal in the fitness function, enabling sparsity to actively compete with performance objectives. It creates a dual competition-attraction mechanism that leverages sparsity-induced signals to create natural parameter competition and repulsion patterns, where pruned regions in one parent model attract complementary parameters from other parents, reducing destructive interference.
    \item We demonstrate its effectiveness using comprehensive empirical evaluation on large-scale LLM benchmarks with multiple architectural scales, demonstrating consistent improvements over strong baselines like particle swarm optimization while maintaining orthogonality with existing merging approaches.

\end{itemize}

\section{Method}


\subsection{Evolutionary Model Merging}

We adopt the main framework from \cite{Abrantes2025}, and define the set of all possible merged models, $\Theta_{\mathcal{M}}$, as: 
\vspace{-3mm}

\begin{equation}
  \Theta_{\mathcal{M}}
  = \left\{ \theta_{\mathcal{M}} \,\middle|\, \theta_{\mathcal{M}} = \mathcal{M}_{\lambda_r}\!\left(\theta_{1}, \ldots, \theta_{K}\right) \right\}
  \label{eq:definition_theta}
  \end{equation}
where $\{\theta_k\}_{k=1}^K$ denotes a set of candidate models to be merged, and $\mathcal{M}_{\lambda_r}$ is a model merging operator parameterized by a mixing ratio $\lambda_r$ that controls the relative contribution of each parent model.
Recent works have explored increasingly expressive merging operators $\mathcal{M}_{\lambda_r}$ to enlarge the capacity of $\Theta_{\mathcal{M}}$~\citep{li2023deep,yang2024model,Abrantes2025}.
Among these, \citet{Abrantes2025} introduces an evolutionary model merging framework that enables a highly flexible parameterization of $\mathcal{M}$.

Rather than directly optimizing the mixing ratio $\lambda_r$ in a continuous manner, the search over $\Theta_{\mathcal{M}}$ is realized through a population-based evolutionary process.
Starting from an initial population of $K$ candidate models, each dense model is expanded by generating multiple sparse variants through pruning operations, resulting in a mixed population containing both dense and sparse individuals.
At each evolutionary step, models in the population are randomly paired to form $\frac{K}{2}$ pairs, and each pair produces one merged offspring.
The offspring models are evaluated and compared against the current population; an offspring replaces an existing individual if it achieves a higher evaluation score.
Through this iterative pairing, evaluation, and replacement process, the population progressively explores the model space induced by $\mathcal{M}_{\lambda_r}$.

Formally, this evolutionary process aims to identify the best-performing merged model
\begin{equation}
  \theta_{\mathcal{M}}^* = \mathcal{M}_{\mathbf{\lambda_r}^*}(\theta_1, \ldots, \theta_K),
\end{equation}
where the optimal parameters $\mathbf{\lambda_r}^*$ are implicitly determined by maximizing the evaluation score
\begin{align}
{\lambda_r}^* = \argmax_{\mathbf{\lambda_r}} 
\sum_{j=1}^{N} 
\mathcal{S}\big(x_j \mid \mathcal{M}_{\mathbf{\lambda_r}}(\theta_1, \ldots, \theta_K)\big).
\end{align}
Here, $\mathcal{S}$ denotes the evaluation score function (e.g., benchmark performance), $x_j$ is a task example, and $N$ is the number of evaluation instances.

Following \cite{Abrantes2025}, the merged model $\theta_{\mathcal{M}}$ is constructed in a layer-wise manner.
Specifically, given two parent models $\theta_A$ and $\theta_B$, the parameters of the merged model are defined as
\begin{equation}
\theta_{\mathcal{M}}^{(l)}
=
\lambda_r^{(l)} \, \theta_A^{(l)}
+
\big(1 - \lambda_r^{(l)}\big) \, \theta_B^{(l)},
\label{eq:layerwise_merge}
\end{equation}
where $\lambda_r^{(l)} \in [0,1]$ denotes a layer-wise instantiation of the mixing ratio, controlling the relative contribution of $\theta_A^{(l)}$ and $\theta_B^{(l)}$.
In contrast to split-point--based formulations, the split parameter $\lambda_s$ is implicitly absorbed by treating each parameter tensor as an independent merging unit.

While \cite{Abrantes2025} also perform layer-wise merging, our method further makes the mixing ratios sparsity-aware by blending evaluation scores with layer-wise sparsity-induced signals.
In our implementation, the layer-wise mixing ratio is computed as
\begin{equation}
\lambda_r^{(l)}
=
\frac{s_A + \omega_A^{(l)}}{(s_A + \omega_A^{(l)}) + (s_B + \omega_B^{(l)})},
\label{eq:score_aware_lambda}
\end{equation}
where $s_A$ and $s_B$ denote the evaluation scores of models $\theta_A$ and $\theta_B$, respectively, and $\omega_A^{(l)}$, $\omega_B^{(l)}$ are layer-wise sparsity-induced weights defined in Section~\ref{sparsity}.

The optimization problem is then reformulated as a search for the best-performing model $\theta_{\mathcal{M}}^*$ exclusively within this subspace $\Theta_{\mathcal{M}}$. 
While the perspective in \cite{Abrantes2025} underscores the role of the merging function in defining the boundaries of the search and constraining the solution space, we exploit the role of the score function $\mathcal{S}$. 
As shown in Figure \ref{fig:merge}, the evolutionary algorithm jointly considers the sparsity-inducing and the task-related objectives for model merging.  

\subsection{Competing for Sparsity}\label{sparsity}

Inspired by the effectiveness of sparsity mechanisms to reduce overfitting, we design a sparsity-inducing process to search for a sparse $\theta_{\mathcal{M}}^*$, which modifies the score function to include sparsity conditions. 
Concretely, $\mathcal{S}(\cdot, \cdot)$ takes in two inputs: a measure of sparsity\footnote{We use the magnitudes of parameters as an indicator, which is most effective and commonly used in the literature~\cite{DSD}. } of model parameters $\theta$,  and the performance measure of models following \cite{Abrantes2025}. 
As shown in Figure \ref{fig:merge}, this is different from sparsifying the merged model iteratively in a subtle but profound way: if the dense network is pruned seprately after merging, the resulting sparsity does not compete with other score factors directly -- that is, changing the sparsity ratio would not change the fitness directly; in contrast, if we incorporate the sparsity into the score function, the sparsity competes with other score factors (e.g., fitness), which introduces more interplays among factors over the whole evolutionary process. 
Our approach also operates in a more fine-grained manner in the sense that the score functions take into account local neural patches, so the sparsity is considered not just on a global level across all parameters. 

Furthermore, consider this scenario: many parameters of $\theta_A$ have been pruned, such that even though $\theta_A$ might have a high score on its utility, $\theta_B$ might occupy slightly more positions because it is not too sparse. 
Also imagine that a subset of weight matrix in $\theta_A$ are more sparse than a subset of weight matrix in $\theta_B$.
In order for weight matrix from $\theta_B$ to take up more resources in the merged model $\theta_{\mathcal{M}}$, the other utility of $\theta_B$ needs to be significantly higher than that of $\theta_A$, which adds extra pressure for the competition between $\theta_A$ and $\theta_B$.  
These examples illustrate how our sparsity-inducing mechanism influences the merging process. 
Our design can be seen as a special form of dense-sparse-dense \cite{DSD}, where our re-dense operation is not based on random initialization, but relies on parents. 
Our iterative dense-sparse mechanism is tailored to model merging tasks, where we should avoid introducing cold-start initialization as we aim not to involve pre- or post-training on large-scale data.   
This sparsity-inducing mechanism to mitigate overfitting issues, which promotes the survival of larger magnitudes of parameters. 

\subsection{Sparsity-Induced Attraction}\label{attraction}


Interestingly, our sparsity-inducing mechanism creates a natural attraction.
If $\theta_A$ is more sparse than $\theta_B$, some non-zero elements of $\theta_B$ would occupy the corresponding zero elements of $\theta_A$. 
We can view this as the zero elements of $\theta_A$, \textit{attracting} the non-zero elements of $\theta_B$. 

\subsection{Annealing Sparsification\label{sec:annealing}}

We propose a simple way to escape from a local suboptimal solution based on joint sparsification. 
Specifically, we anneal the sparsification ratio of the merged model during training, which can be seen as the mutation operator and is inspired by \cite{1608.03983}.

Note that we apply this sparsification annealing schedule to both the individual models and the merged model. 
This encourages our pipeline to explore the merged model space, especially in the early stage more effectively. 







\section{Experiments}
We conduct our experiments based on fine-tuned variants of LLaMA-3 models\footnote{We exclude Qwen2.5 models as Qwen2.5-Coder is continually trained with substantially more tokens than its base counterpart~\citep{hui2024qwen25coder}, leading to reduced compatibility for parameter-space merging.
Although recent work reports merging Qwen2.5-Coder and Qwen2.5-Instruct at the 7B scale~\citep{sigrist2025pipelineassessmergingmethods}, it focuses on task-vector or interpolation-based methods and does not consider iterative evolution-based merging, which is the focus of our study.}, each with 3 billion parameters, which are pretrained from the same base architecture but specialized for different competencies: mathematical reasoning and multilingual understanding, respectively. 
These models are sourced from the MergeBench benchmark suite~\citep{tang2024fusionbench}.

Specifically, we use the following models as merging candidates:
\begin{itemize}
    \item LLaMA-3.2-3B-Instruct-Math\footnote{\url{https://www.llama.com/models/llama-3/}}, specialized for mathematical reasoning tasks
	\item 	LLaMA-3.2-3B-Instruct-Multilingual, specialized for multilingual understanding and reasoning
\end{itemize}



To perform model merging optimization, we randomly sample each time 1,319 instances from GSM8K training set and 200 instances from MMLU-ProX training set as dynamic optimization set. We evaluate the merged models on the full GSM8K test set and on a subset of 1,000 instances from the MMLU-ProX test set. 

For MMLU-ProX, the 1,000 evaluation samples are stratified by language, ensuring that the subset preserves the original language distribution of the full test set, and we use the full test sets of GSM8K to assess the generalization performance of the merged models.

As our primary baseline, we compare against particle swarm optimization (PSO), a strong and recently proposed evolutionary approach for parameter-space model fusion. 
All experimental results are reported with the identical evaluation protocols to ensure fair comparison.


\subsection{Main Results}

We first compare SAE against strong baselines on multiple benchmarks.
Table~\ref{tab:main} reports the main results.
SAE slightly but consistently outperforms PSO on both GSM8K and MMLU-ProX,
indicating that sparsity-aware archive-based optimization provides a more
effective exploration of the parameter merging space.

\begin{table}[t!]
\centering
\footnotesize
\begin{tabular}{l ccc}
\toprule
\multirow{2}{*}{\textbf{Method}} 
& \multicolumn{3}{c}{\textbf{Math + Multilingual}} \\
\cmidrule(lr){2-4}
& \textbf{GSM8K} & \textbf{MMLU-ProX} & \textbf{Avg.} \\
\midrule
Task Arithmetic & 0.741 & 0.187 & 0.464 \\
Weight Average & 0.742 & 0.185 & 0.464 \\
Rankmean & 0.137 & 0.176 & 0.157 \\
PSO & 0.7801 & 0.164 & 0.472 \\
SAE (Global) & \textbf{0.798} & 0.170 & \textbf{0.484} \\
SAE (Local) & 0.7748 & \textbf{0.182} & 0.478 \\
\bottomrule
\end{tabular}
\caption{Comparing SAE with strong baselines under the Math + Multilingual fusion setting. Task arithmetic is from \citep{ilharco2023}, and other baselines can be refereed to \citep{Zhang2025}.}
\label{tab:main}

\end{table}


\begin{figure*}[h!]
  \centering
  \setlength{\tabcolsep}{3pt}
  \begin{tabular}{cccc}
    \includegraphics[width=0.24\textwidth]{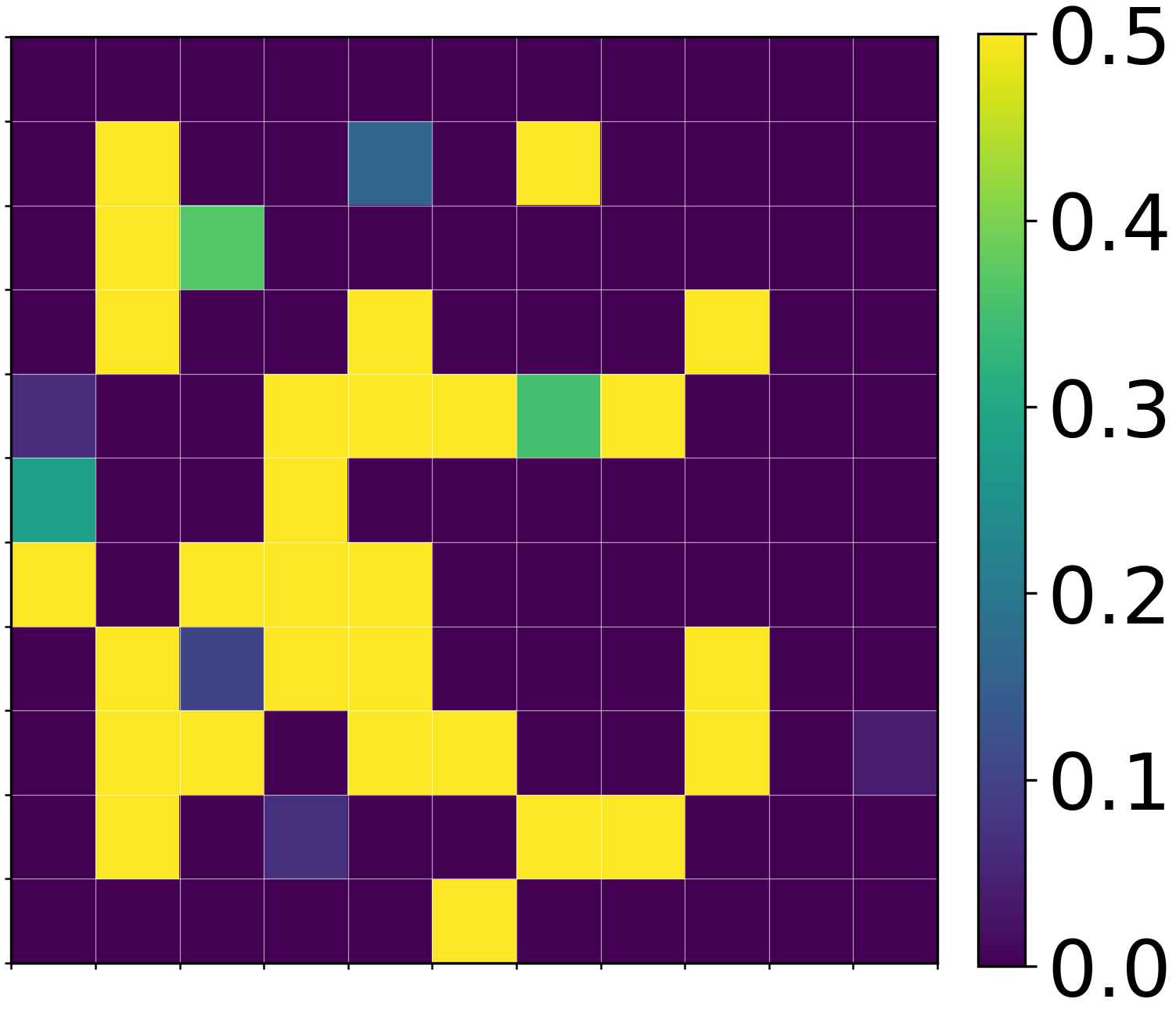} &
    \includegraphics[width=0.24\textwidth]{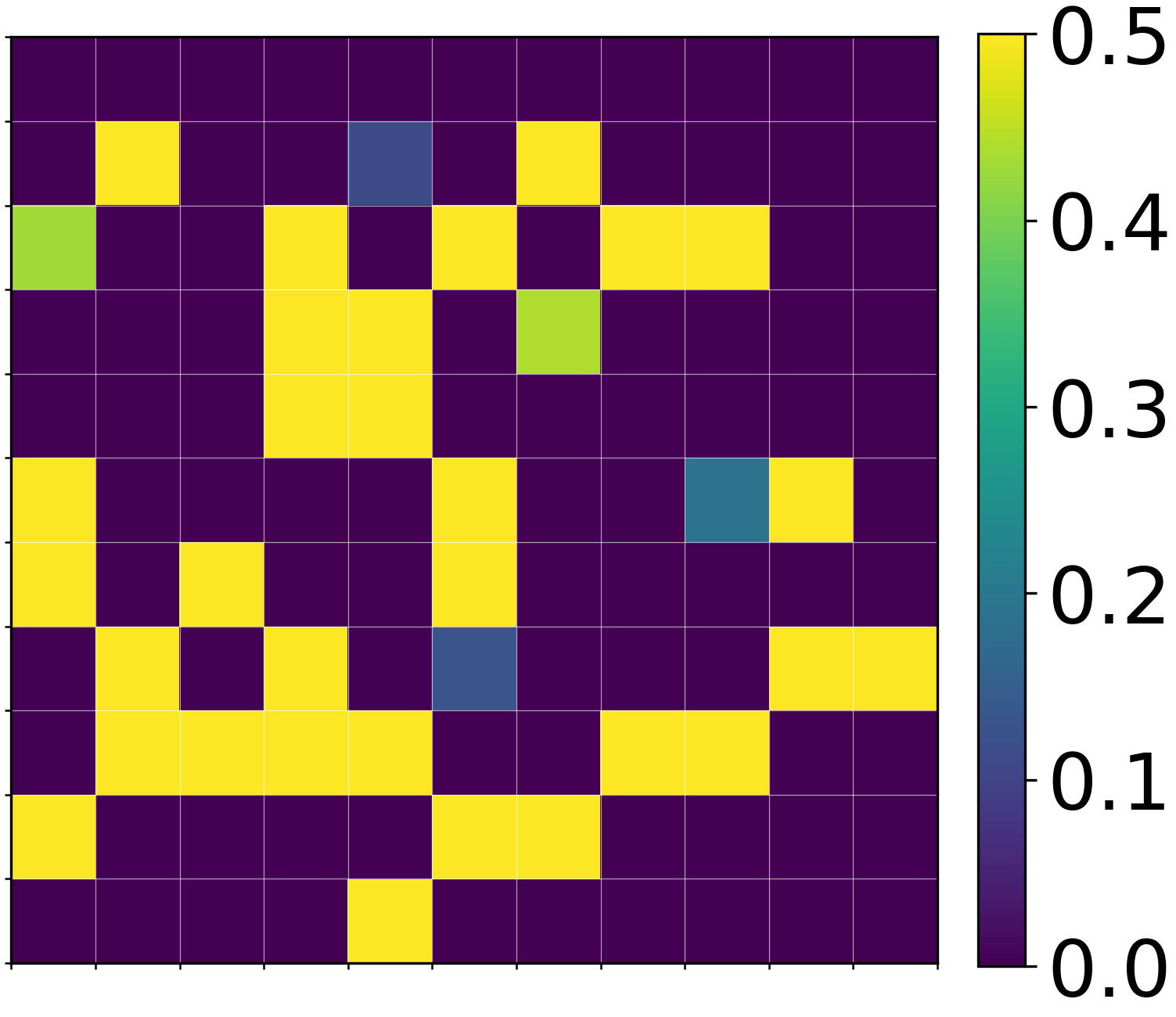} &
    \includegraphics[width=0.24\textwidth]{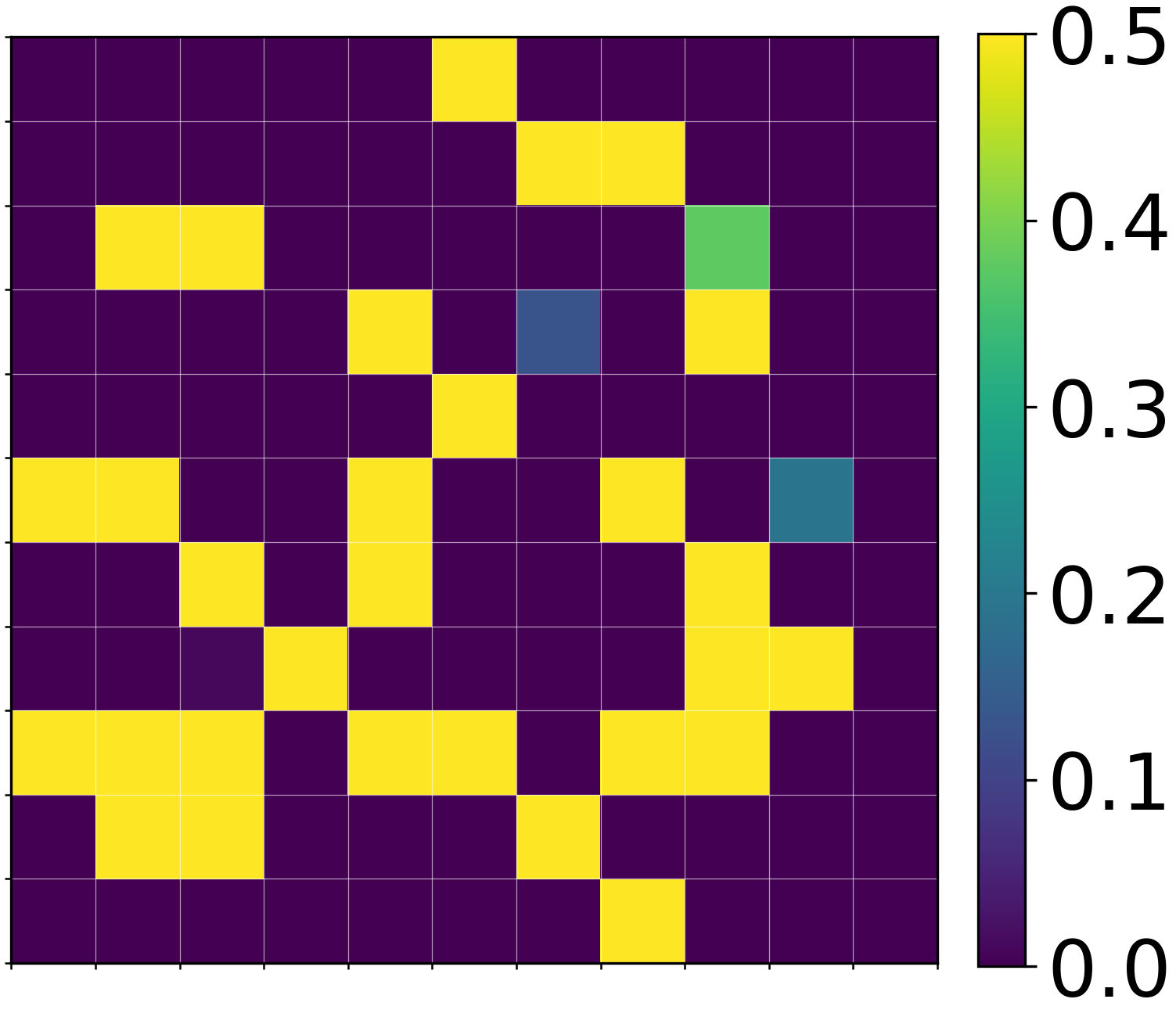} &
    \includegraphics[width=0.24\textwidth]{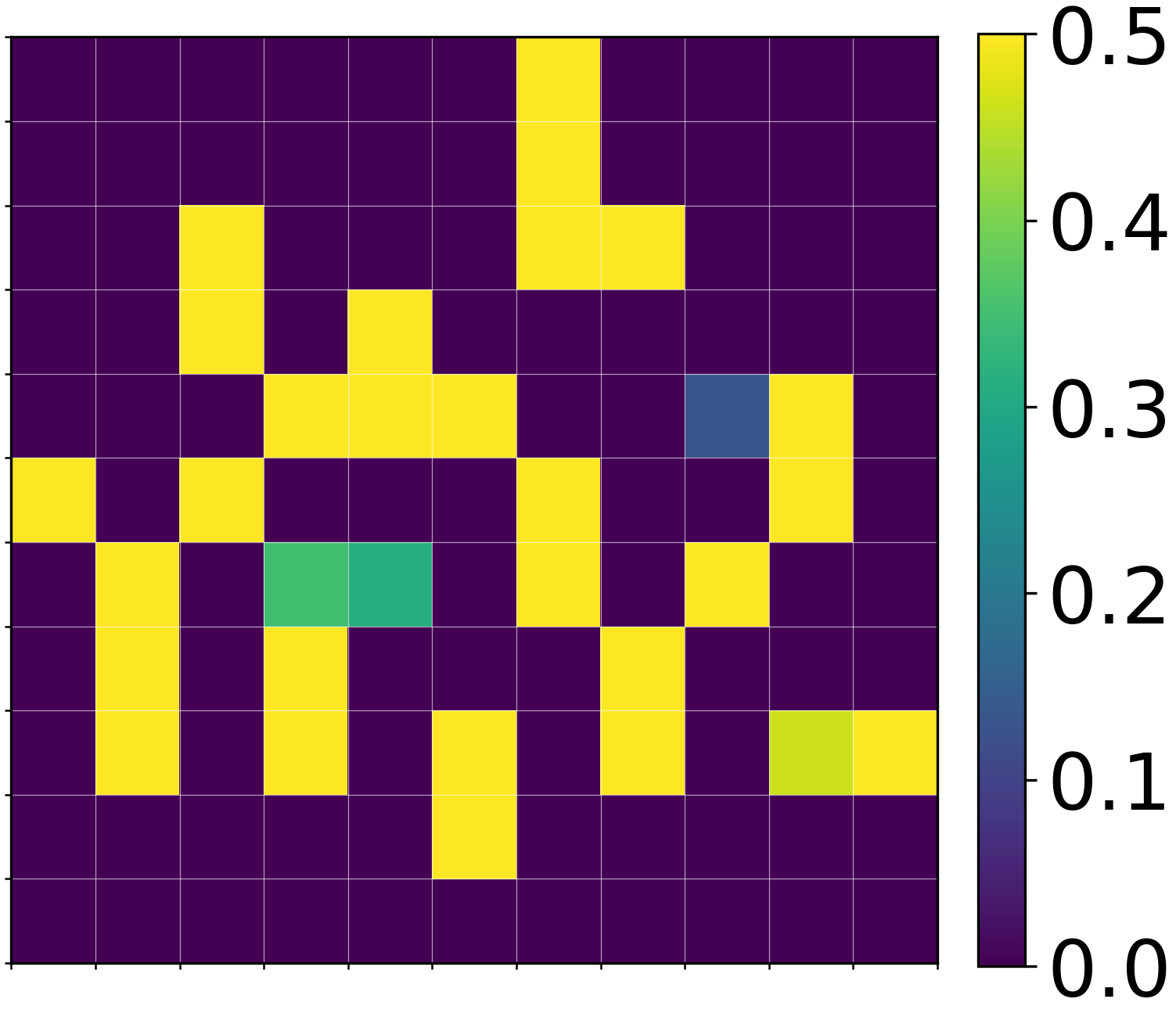} \\
    (a) Math expert &
    (b) Multilingual expert &
    (c) PSO-merged &
    (d) SAE-merged \\
  \end{tabular}
  \caption{\textbf{Convexity landscapes on MMLU-ProX.} Each cell corresponds to a parameter point
  $\theta(\alpha,\beta)=\theta_0+\alpha d_1+\beta d_2$ along two random directions (layer-wise normalized),
  colored by a local convexity score computed from Hessian spectra:
  \texttt{convexity = abs(lambda\_min) / (abs(lambda\_max) + eps)} (clipped to \texttt{[0, 0.5]}).
  Brighter regions indicate more balanced positive/negative curvature (i.e., relatively stronger non-convexity),
  while darker regions indicate one-sided curvature dominance.}
  \label{fig:convexity_mmlu_prox}
\end{figure*}

\begin{figure*}[t!]
\centering
\begin{tabular}{cccc}

\includegraphics[width=0.23\textwidth]{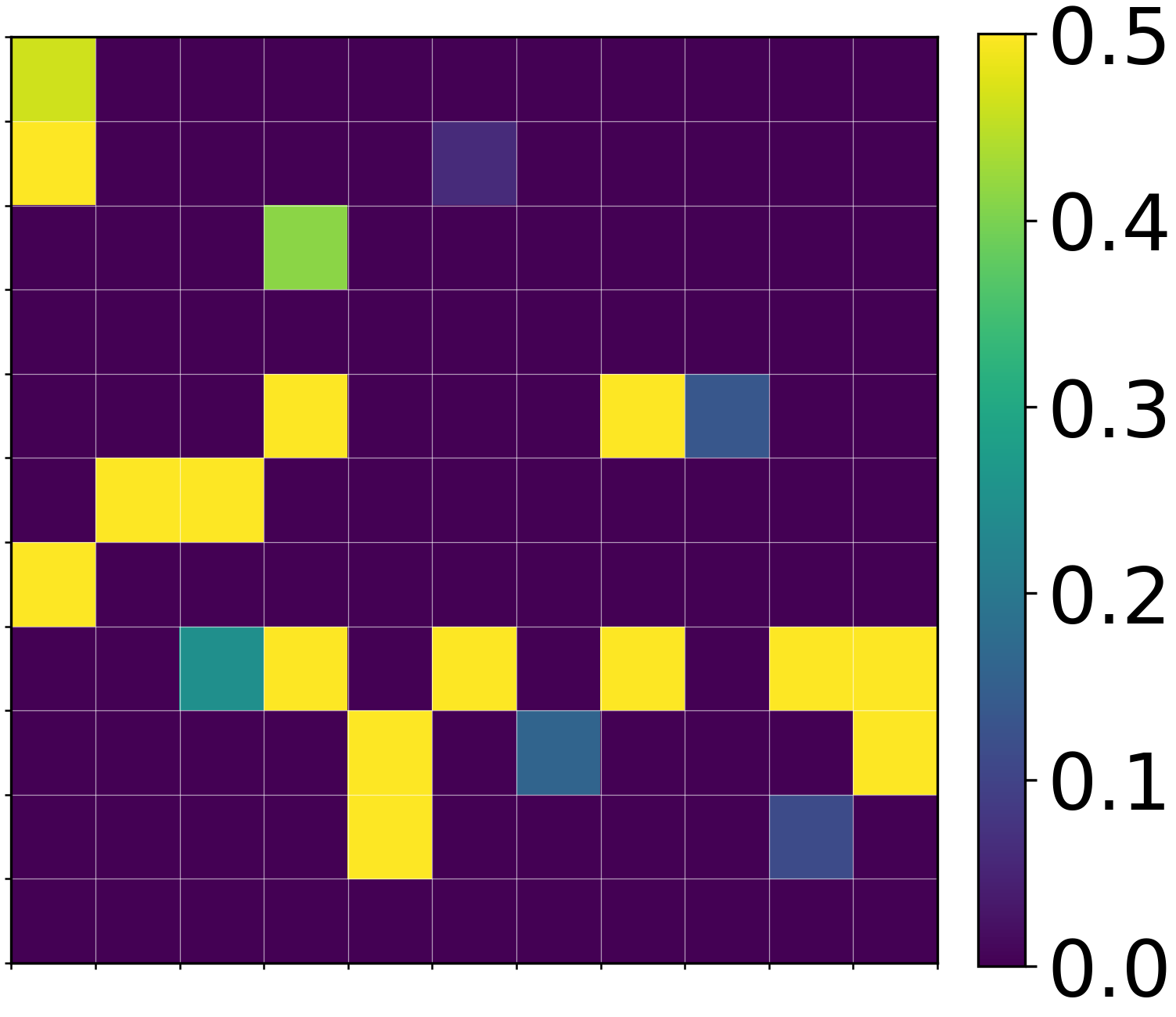} &
\includegraphics[width=0.23\textwidth]{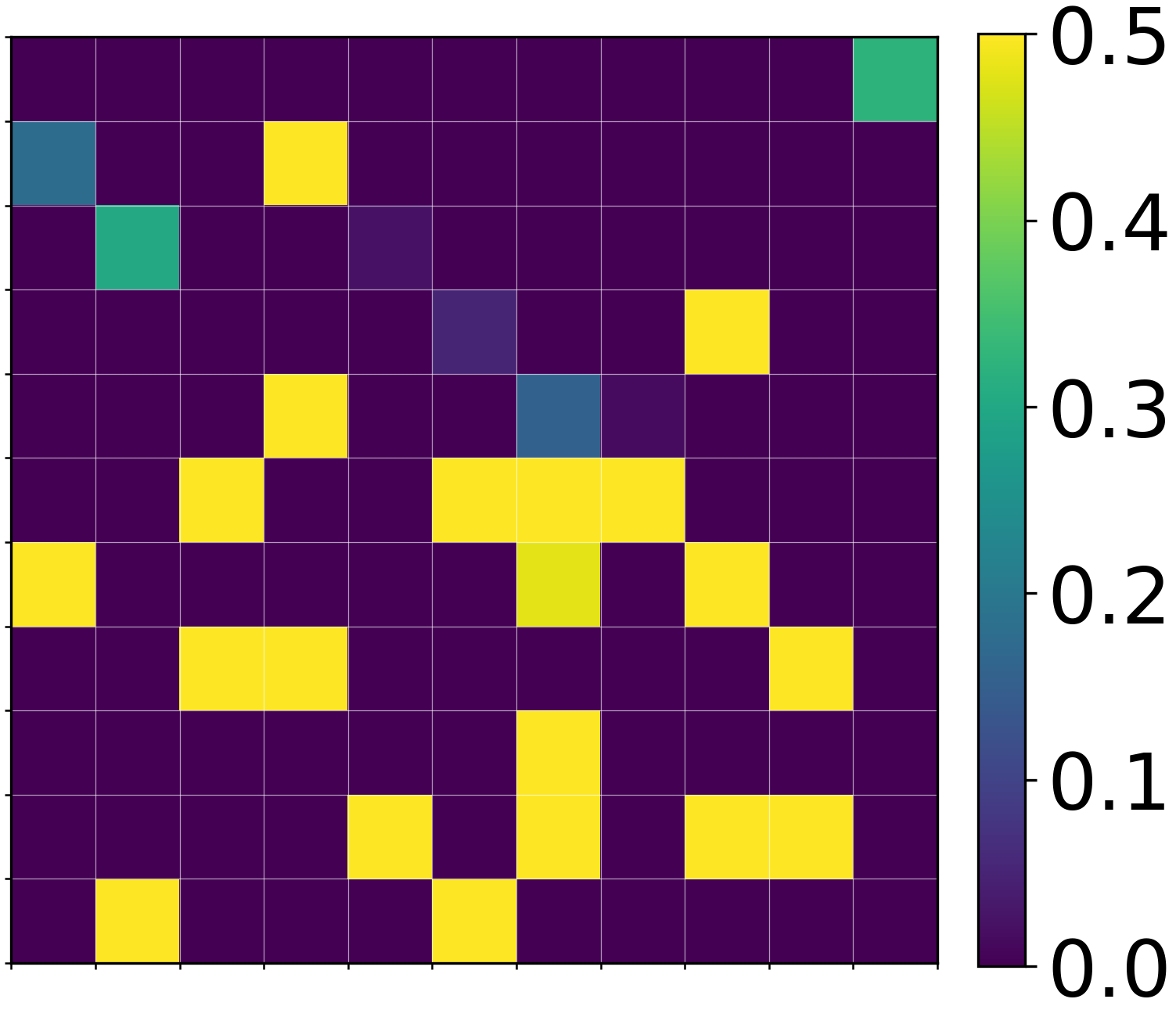} &
\includegraphics[width=0.23\textwidth]{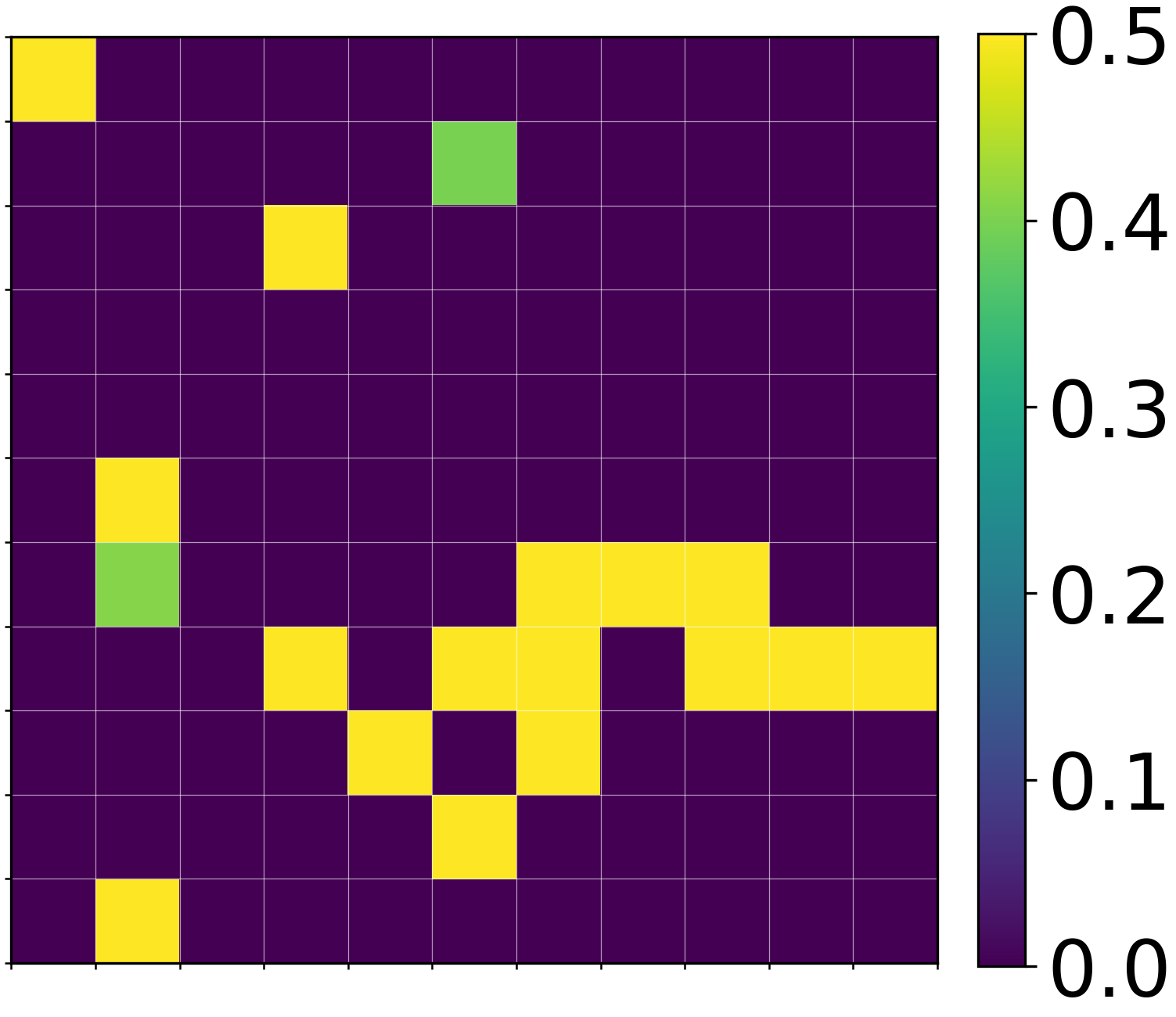} &
\includegraphics[width=0.23\textwidth]{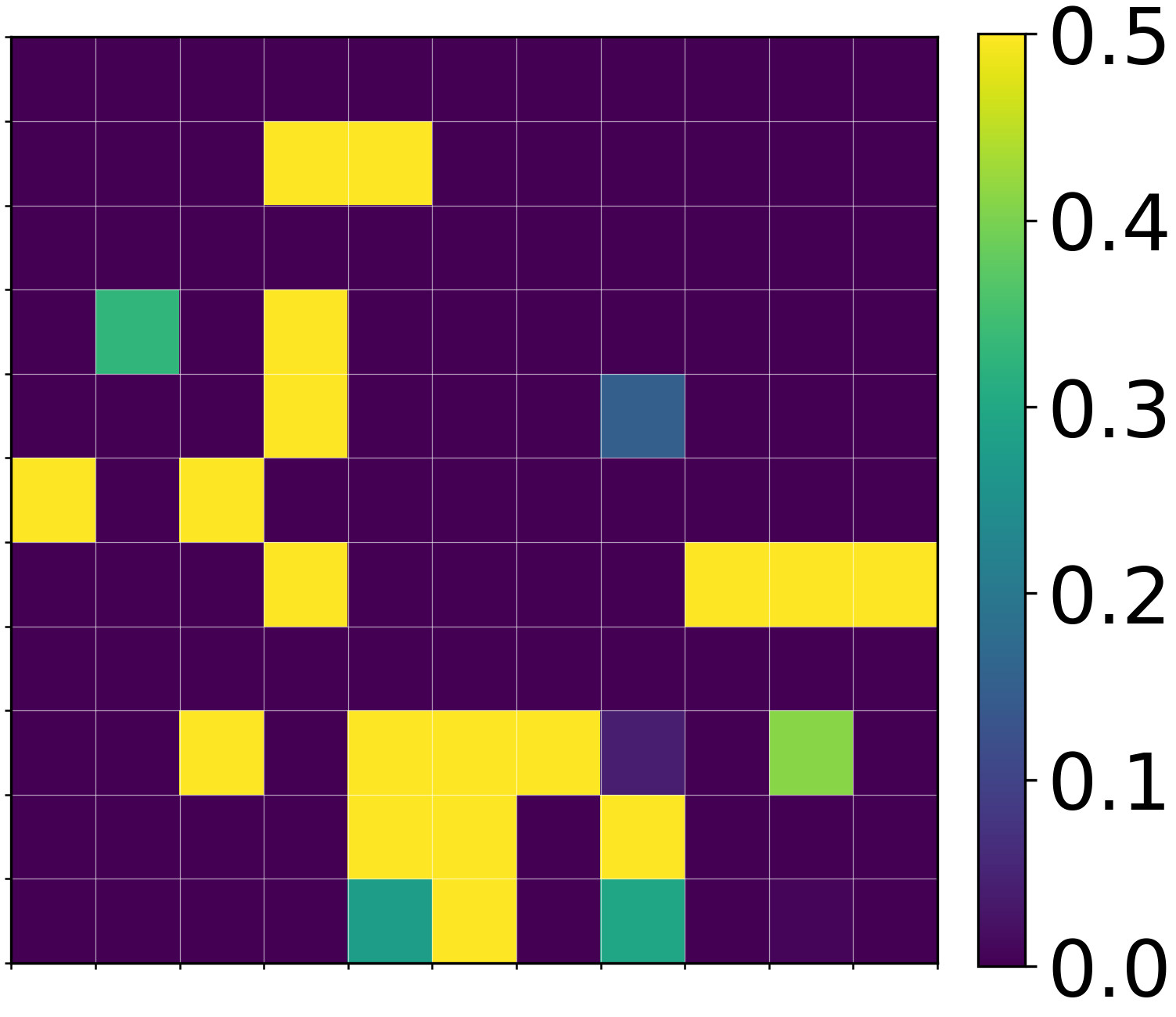} \\

\small (a) Math expert &
\small (b) Multilingual expert &
\small (c) PSO-merged &
\small (d) SAE-merged

\end{tabular}

\vspace{1mm}
\caption{\textbf{Convexity landscapes on GSM8K.}
Each cell corresponds to a parameter point
$\theta(\alpha,\beta)=\theta_0+\alpha d_1+\beta d_2$
along two shared random directions (layer-wise normalized).
Cells are colored by a Hessian-based convexity proxy
computed from the extreme eigenvalues:
\texttt{convexity = abs(lambda\_min) / (abs(lambda\_max) + eps)},
clipped to $[0, 0.5]$.
Brighter regions indicate more balanced positive/negative curvature,
while darker regions indicate one-sided curvature dominance.
}
\label{fig:convexity_gsm8k}
\vspace{-3mm}
\end{figure*}

In addition to the quantitative results, we also analyze the geometric properties of the merged solutions by visualizing their loss
landscapes along shared random directions, following the methodology of
\citet{li2018visualizinglosslandscapeneural}.
Figure~\ref{fig:convexity_mmlu_prox} further visualizes the local geometry
on MMLU-ProX using a Hessian-based convexity proxy: the two base experts exhibit
mostly isolated high-convexity (bright) cells, suggesting highly localized
curvature imbalance, whereas PSO produces more scattered bright patches without
clear continuity. In contrast, the SAE-merged model shows more contiguous and
structurally coherent regions in the convexity map, suggesting a smoother and
more consistent second-order landscape after sparsity-aware optimization.

We report convexity visualizations on GSM8K in
Figure~\ref{fig:convexity_gsm8k}, following the same protocol as the main-text
analysis on MMLU-ProX.

Both the math expert and the multilingual expert exhibit mostly isolated
high-convexity (bright) cells, indicating localized curvature imbalance along
the sampled directions.
Compared to the expert models, PSO introduces additional high-convexity regions,
but these regions remain spatially scattered and lack clear structural
continuity.
In contrast, the SAE-merged model shows relatively more contiguous and locally
coherent convexity patterns, with fewer isolated extrema.

Although the overall loss geometry on GSM8K appears smoother than on MMLU-ProX,
these observations are qualitatively consistent with the main-text results.
Together with the loss landscape visualizations in
Figure~\ref{fig:loss_landscape_sae_vs_pso}, these results suggest that
sparsity-aware optimization consistently regularizes the local second-order
geometry across tasks, rather than merely inheriting expert-specific curvature
structures.

\subsection{Ablation Study}
We analyze how different design choices affect performance.
Results are summarized in Table~\ref{tab:ablation}.

\begin{table*}[t]
\centering
\small
\begin{tabular}{l l c c c}
\toprule
\textbf{Setting} 
& \textbf{Configuration} 
& \textbf{GSM8K} 
& \textbf{MMLU-ProX} 
& \textbf{Avg.} \\
\midrule

SAE (Default)
& $\mathcal{S}_{\text{global}},\ s\!\in[0.1,0.6],\ T_0{=}3,\ T_{\text{mult}}{=}2$
& 0.7824 & 0.1680 & 0.4752 \\

Redense with original dense
& $\mathcal{S}_{\text{global}},\ \theta \leftarrow \theta_{\text{dense}}^{(0)}$
& \textbf{0.7915} & 0.1610 & 0.4763 \\

Local layer-wise sparsity
& $\mathcal{S}_{\text{local}},\ s\!\in[0.1,0.6]$
& 0.7748 & \textbf{0.1810} & 0.4779 \\

Larger sparse-rate range
& $\mathcal{S}_{\text{global}},\ s\!\in[0.05,0.9]$
& \textbf{0.7983} & \textbf{0.1700} & \textbf{0.4842} \\

Slower schedule
& $\mathcal{S}_{\text{global}},\ T_0{=}4$
& 0.7862 & 0.1520 & 0.4691 \\

\bottomrule
\end{tabular}
\caption{Ablation results of SAE under different design choices.
Unless otherwise specified, all settings use global sparsity scoring and the default cyclic sparsity schedule.
$\mathcal{S}_{\text{global}}$ and $\mathcal{S}_{\text{local}}$ denote global and layer-wise sparsity scoring strategies, respectively.
\textbf{Redense with original dense} denotes a variant where the re-densification phase initializes parameters from the original dense model $\theta_{\text{dense}}^{(0)}$, rather than inheriting weights from the current parent models.}
\label{tab:ablation}
\end{table*}

The ablation study reveals that increasing the sparsity-rate search range
consistently improves performance on both tasks.
Layer-wise sparsity benefits multilingual reasoning but degrades mathematical
accuracy, suggesting task-dependent sensitivity to sparsity granularity.
Slower sparsity annealing fails to provide further gains.

\subsection{Effect of Archive Size}

Table~\ref{tab:pop} reports the impact of archive population size on SAE and PSO.
Increasing the archive size yields limited benefit for PSO,
while SAE shows clear improvement on MMLU-ProX as the population grows.
This suggests that archive diversity plays a critical role in sparsity-aware
merging, particularly for multilingual reasoning.

\begin{table}[H]
\centering
\small
\begin{tabular}{lcc}
\toprule
\textbf{Method} & \textbf{GSM8K} & \textbf{MMLU-ProX}  \\
\midrule
SAE (pop = 8, default) & 0.7824 & 0.1680 \\
SAE (pop = 16) & 0.7688 & 0.1700 \\
SAE (pop = 32) & \textbf{0.7847} & \textbf{0.1790} \\
PSO (pop = 8) & 0.7801 & 0.1640 \\
PSO (pop = 16) & 0.7847 & 0.1670 \\
PSO (pop = 32) & 0.7817 & 0.1610 \\
\bottomrule
\end{tabular}
\caption{Impact of archive population size on SAE and PSO.}
\label{tab:pop}
\vspace{-3mm}
\end{table}

\subsection{Cyclic Sparsity Hyperparameter Analysis}
\vspace{-3mm}
\begin{table}[H]
\centering
\small
\begin{tabular}{ccccc}
\toprule
$s_{\min}$ & $s_{\max}$ & GSM8K & MMLU-ProX & Remark \\
\midrule
0.10 & 0.60 & 0.7824 & 0.1680 & Default \\
0.10 & 0.70 & 0.7703 & 0.1670 & Comparable \\
0.05 & 0.90 & \textbf{0.7983} & \textbf{0.1700} & Best \\
\bottomrule
\end{tabular}
\caption{Effect of sparsity-rate range on SAE performance.}
\label{tab:srange}
\vspace{-3mm}
\end{table}

Table~\ref{tab:srange} summarizes the effect of different sparsity-rate ranges on SAE performance.
Expanding the sparsity-rate range significantly enhances performance,
indicating that broader exploration of sparsity configurations
is crucial for effective model merging.

\subsection{Sparse Measurement Variant}
\vspace{-3mm}
\begin{table}[H]
\centering
\small
\begin{tabular}{lcc}
\toprule
\textbf{Method} & \textbf{GSM8K} & \textbf{MMLU-ProX} \\
\midrule
SAE (magnitude, default) & 0.7824 & 0.1680 \\
SAE (zero-count) & \textbf{0.7832} & \textbf{0.1750} \\
PSO (baseline) & 0.7801 & 0.1640 \\
\bottomrule
\end{tabular}
\caption{Comparison of different sparsity measurement strategies.}
\label{tab:sparse}
\vspace{-3mm}
\end{table}

As shown in Table~\ref{tab:sparse}, using zero-count as the sparsity measure further improves performance,
suggesting that explicit structural sparsity better correlates
with downstream task accuracy.

\subsection{Cyclic Sparsity Scheduling}

We adopt a cyclic scheduling strategy for sparsity control inspired by the restart mechanism of SGDR.
Unlike classical SGDR, which operates on gradient-based learning rates, our method does not involve gradients or optimizer dynamics.
Instead, we treat the sparsity rate as an explicit control variable and apply cyclic modulation directly to the sparse ratio during training.

Specifically, the sparsity rate is scheduled to increase from
$s_{\min}$ to $s_{\max}$ within each cycle, followed by a restart that resets the sparsity to a lower value.
The cycle length is initialized as $T_0$ and expanded multiplicatively by a factor of $T_{\text{mult}}$ after each restart.
This design induces alternating phases of strong structural constraint (high sparsity) and relaxed exploration (low sparsity), enabling the model to balance structural consolidation and re-exploration over time.

We interpret sparsity as a form of architectural regularization rather than an optimization hyperparameter.
High sparsity enforces compact and stable subnetworks, while lower sparsity allows broader parameter exploration. Cyclic modulation of sparsity therefore plays a role analogous to annealing or curriculum strategies, but operates at the level of network structure instead of gradient dynamics.

\subsection{Cyclic Sparsity Schedule Ablation}

\begin{table}[h]
\centering
\small
\setlength{\tabcolsep}{3pt}
\begin{tabular}{cccccc}
\toprule
$T_0$ & $T_{\text{mult}}$ & \textbf{GSM8K} & \textbf{MMLU-ProX} & \textbf{Avg.} & \textbf{Remark} \\
\midrule
3 & 2 & \textbf{0.7824} & 0.1680 & \textbf{0.4752} & Default \\
2 & 2 & 0.7612 & \textbf{0.1830} & 0.4721 & $T_0\downarrow$ \\
2 & 1 & 0.7809 & 0.1630 & 0.4720 & $T_0\downarrow$, no grow \\
3 & 1 & 0.7665 & 0.1710 & 0.4688 & no grow \\
\bottomrule
\end{tabular}
\caption{Cyclic sparsity schedule ablation ($s_{\min}=0.05$, $s_{\max}=0.9$).
``no grow'' denotes $T_{\text{mult}}{=}1$, i.e., no cycle length expansion.}
\label{tab:sgdr_cycle}
\vspace{-3mm}
\end{table}

\paragraph{Default Configuration.}
Unless otherwise specified, we use a fixed default cyclic sparsity configuration throughout all experiments.
Specifically, the sparsity rate is bounded by $s_{\min}=0.1$ and $s_{\max}=0.6$, with a total of $12$ training steps.
The initial cycle length is set to $T_0=3$, and the cycle length is expanded after each restart by a multiplicative factor of $T_{\text{mult}}=2$.
This configuration is treated as the default setting in all subsequent ablation studies.

Table~\ref{tab:sgdr_cycle} presents an ablation study of the cyclic sparsity scheduling parameters.
Compared to the default configuration, reducing the initial cycle length
results in more frequent sparsity resets, which biases the model toward improved multilingual generalization at the cost of mathematical reasoning performance.
In contrast, disabling cycle expansion ($T_{\text{mult}}{=}1$) removes long-term sparsity annealing and generally degrades overall stability.
These results indicate that cyclic sparsity scheduling plays a critical role in balancing task-specific structural exploration and consolidation, independently of gradient-based learning dynamics.

\section{Related Work}



Our work is broadly related to competitive computation mechanisms, such as compute-to-compute~\cite{compete_compute}, where multiple candidates compete and only the winner dominates the output.
Such competition naturally induces sparsity and specialization, but is typically defined at the activation or routing level rather than directly in parameter space. Recent work also exploits sparsity to encourage diversity, for example by generating multiple sparse variants of a model~\cite{Zhang2025}, though the sparsity level in these approaches is usually heuristic and not explicitly optimized.

In contrast, our method treats sparsity as an explicit optimization signal and integrates it directly into the evolutionary objective,
allowing sparsity to actively regulate competition and interaction during model merging.

\paragraph{Model Merging}

Early studies showed that pretrained networks could be combined in weight space to share complementary abilities without joint training.
Model Soup~\cite{Wortsman2022} demonstrated that aggregating fine-tuned checkpoints improves robustness,
while Task Vectors~\cite{ilharco2023} further revealed that fine-tuning updates behave like linear directions in parameter space.
Yet naïve averaging often causes destructive interference and offers little control over conflicting updates.

Building on this foundation, later methods sought more principled ways to stabilize merging.
TIES~\cite{yadav2023} interpolates weights according to task-wise interference scores, and DARE~\cite{66yu2024} down-weights incompatible updates through stochastic sparsification.
Beyond fixed-rule, non-iterative merging, a growing line of work formulates model merging as an evolutionary or population-based search problem.
Early efforts adapt black-box optimizers such as CMA-ES to search over merging coefficients or structures.
More recent approaches emphasize population diversity and interaction:
M2N2~\cite{Abrantes2025} maintains multiple niches through competition and attraction,
while PSO-Merging~\cite{Zhang2025} formulates merging as a particle swarm optimization process.
Other work explores explicit mutation and crossover on model weights~\cite{du2024knowledge},
and reusable frameworks such as Mergenetic~\cite{Minut_2025} and MergeKit~\cite{goddard-etal-2024-arcees} further reflect the rapid development of model merging methods.

Unlike dense-space merging methods such as M2N2, which operationalize competition and attraction using only dense weights and global performance signals,
SAE directly incorporates sparsity into the merging objective.
Sparsity thus acts as both a regulatory signal and a structural mechanism—pruning clears parameter slots that other parents can fill—strengthening fine-grained complementarity, improving exploration, and reducing overfitting relative to dense-only designs.

In the pursuit of efficient merging, sparsity has also been explored for scaling multi-task fusion~\citep{davari2024model}.
However, prior work does not explicitly model the balance between sparsity and competition during merging, which is the focus of our approach.
\vspace{-2mm}

\paragraph{Model Pruning}
Model pruning has long been studied as an effective regularization technique to improve efficiency and generalization.
Classic methods identify important parameters based on magnitude, sensitivity, or training dynamics, including the Lottery Ticket Hypothesis~\cite{frankle2019},
as well as single-shot or data-agnostic approaches such as SNIP~\cite{lee2019} and SynFlow~\cite{tanaka2020}.
More recent work extends pruning to large pretrained models and LLMs, with methods such as SparseGPT~\cite{Frantar2023SparseGPT} and WANDA~\cite{sun2024wanda}
that leverage structured or N:M sparsity patterns for scalable compression.

Beyond single-model efficiency, sparsification has also been explored in model merging.
Sparse Model Soups~\cite{zimmer2024} combines pruning with model averaging,
while pruning-aware merging methods~\cite{he2021pruning,zhu2024} mitigate parameter conflicts in multitask or cross-domain settings.
In evolutionary merging, PSO-Merging~\cite{Zhang2025} applies random sparsification to increase population diversity during initialization.

However, in most existing approaches, sparsity is treated as a preprocessing step or auxiliary heuristic rather than an explicit optimization objective.
In contrast, our approach interleaves sparsification and re-densification with merging throughout the evolutionary process,
treating sparsity as a first-class evolutionary signal that competes with task performance and actively shapes the merging dynamics.

\section{Conclusion}

In this work, we have presented a Sparsity-Aware Evolution (SAE) framework that fundamentally rethinks the role of sparsity in model merging. 
By shifting the paradigm from static parameter averaging to a dynamic, evolutionary search driven by sparsity constraints, we successfully mitigated the destructive interference that typically plagues multi-task fusion. Our results demonstrate that treating sparsity as an active selection pressure—rather than a mere regularizer—forces the emergence of modular, conflict-free subnetworks, thereby allowing the merged model to effectively synthesize the distinct capabilities of its parents through iterative pruning and re-densification. Ultimately, this work establishes that the strategic subtraction of parameters is as vital as their aggregation, offering a scalable and efficient pathway for developing versatile LLMs without the need for extensive retraining.

\section*{Limitations}

While our sparsity-aware evolution framework demonstrates clear gains in merging reliability and modularity, it introduces certain trade-offs compared to simple linear merging techniques. 
First, the evolutionary search process, though more efficient than full retraining, incurs a higher computational cost than one-shot methods like task arithmetic due to the need to evaluate multiple generations of candidate models. 
Second, our current experiments primarily validate the approach on homologous models sharing the same base architecture (LLaMA-3); its efficacy in merging heterogeneous architectures or models with vastly different pre-training distributions remains an open question. 
Third, while the re-dense mechanism effectively repopulates pruned subspaces, the optimal schedule for annealing sparsity is currently heuristic-based, suggesting that future work could benefit from adaptive, meta-learned schedules to further automate the balancing of competition and attraction.
Finally, our proposed pipeline is a general-purpose for LLMs and not specifically designed for MoE models.
We have not tested its effectiveness for MoE models yet, which could be a promising next step.



\bibliography{sparse_merge}

\appendix

\section{Loss Surface Analytics}

\begin{figure*}[t!]
\centering
\begin{tabular}{ccc}

\includegraphics[width=0.31\textwidth]{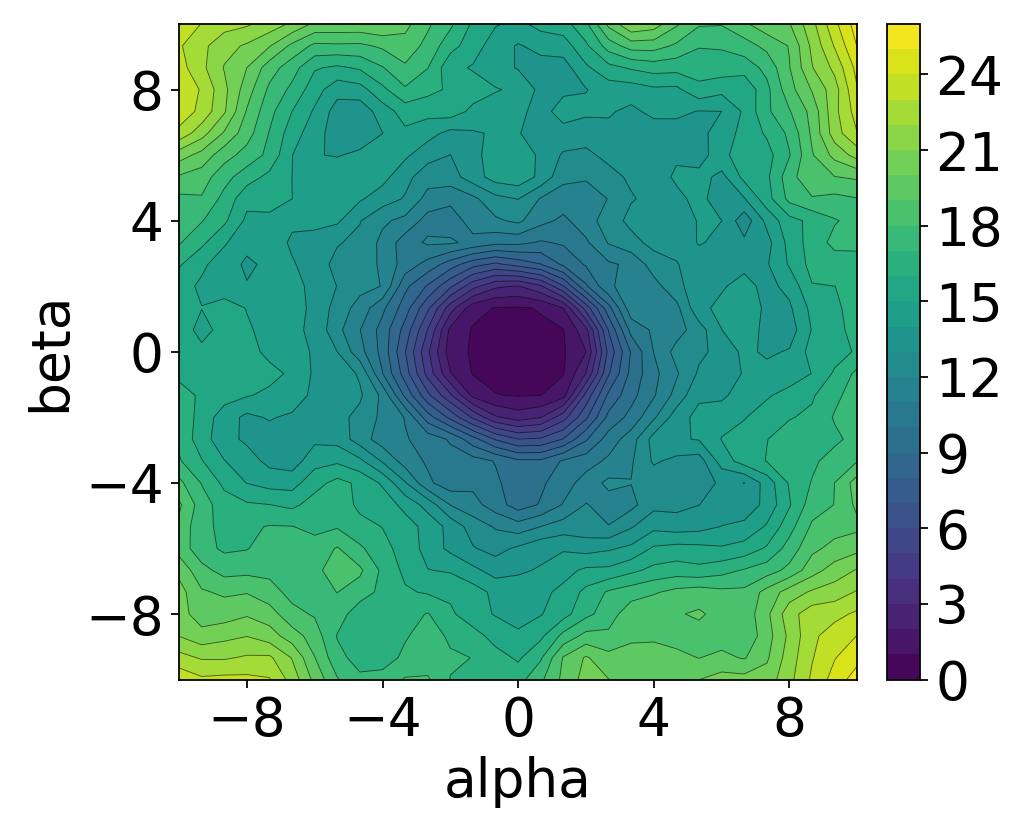} &
\includegraphics[width=0.31\textwidth]{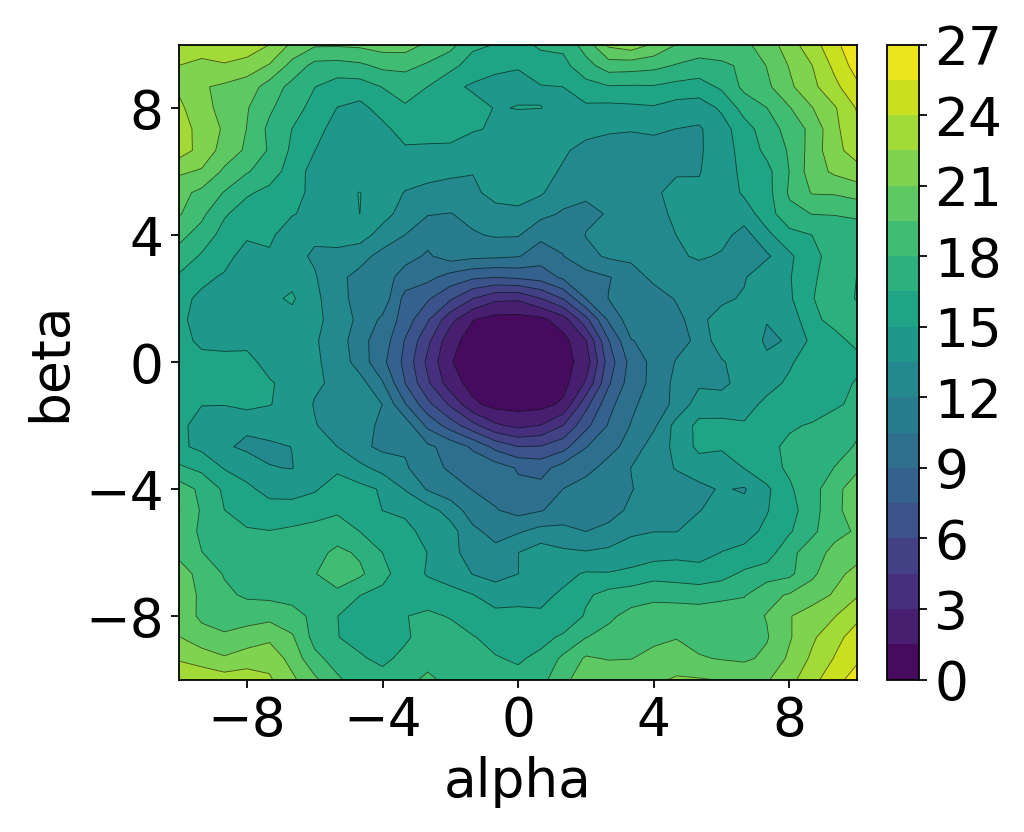} &
\includegraphics[width=0.31\textwidth]{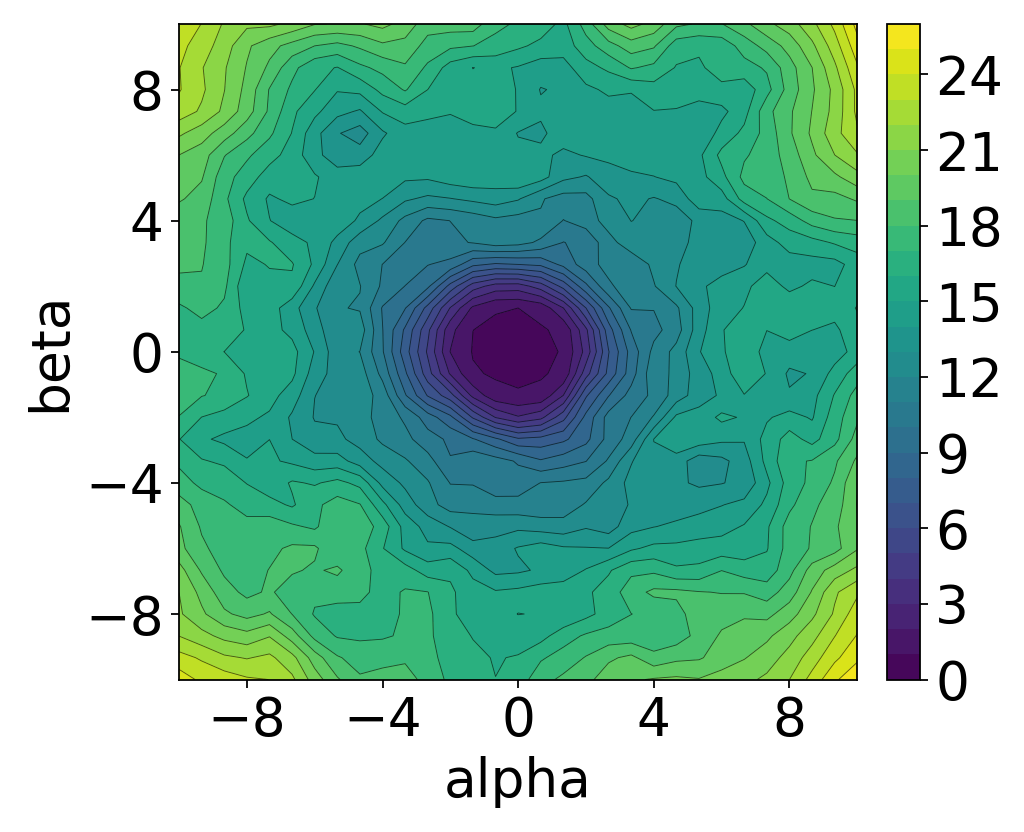} \\
\small Math expert (GSM8K) &
\small PSO (GSM8K) &
\small SAE (GSM8K)  \\[2mm]

\includegraphics[width=0.31\textwidth]{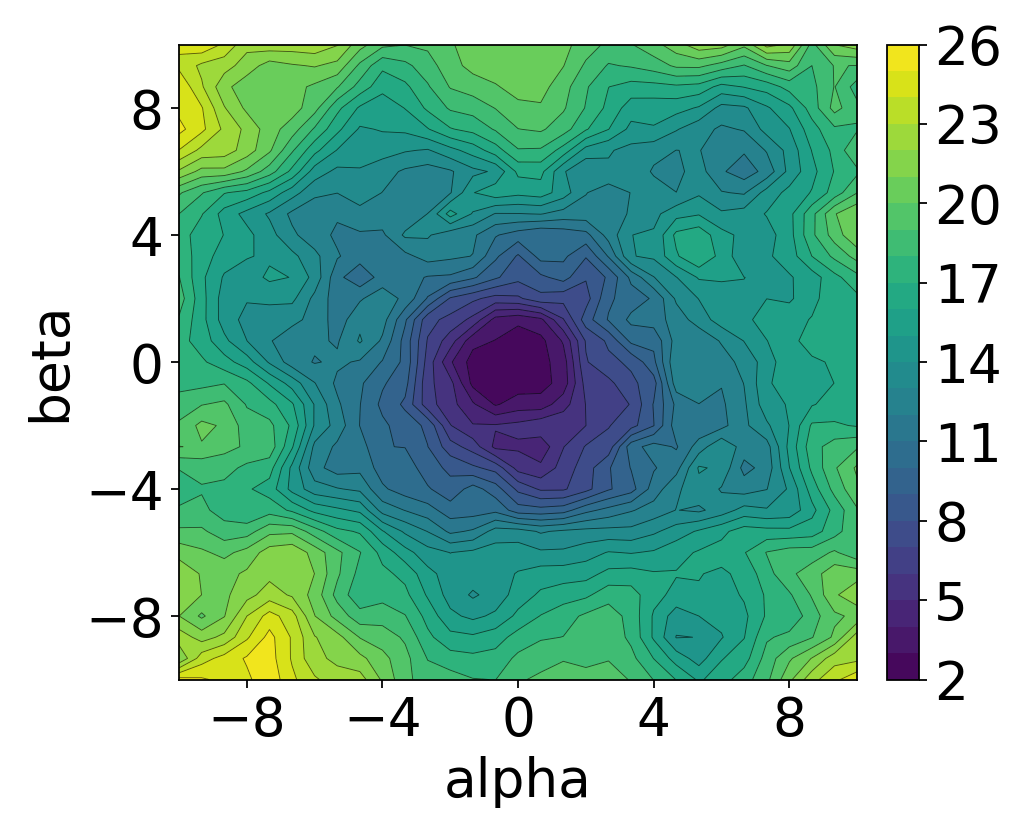} &
\includegraphics[width=0.31\textwidth]{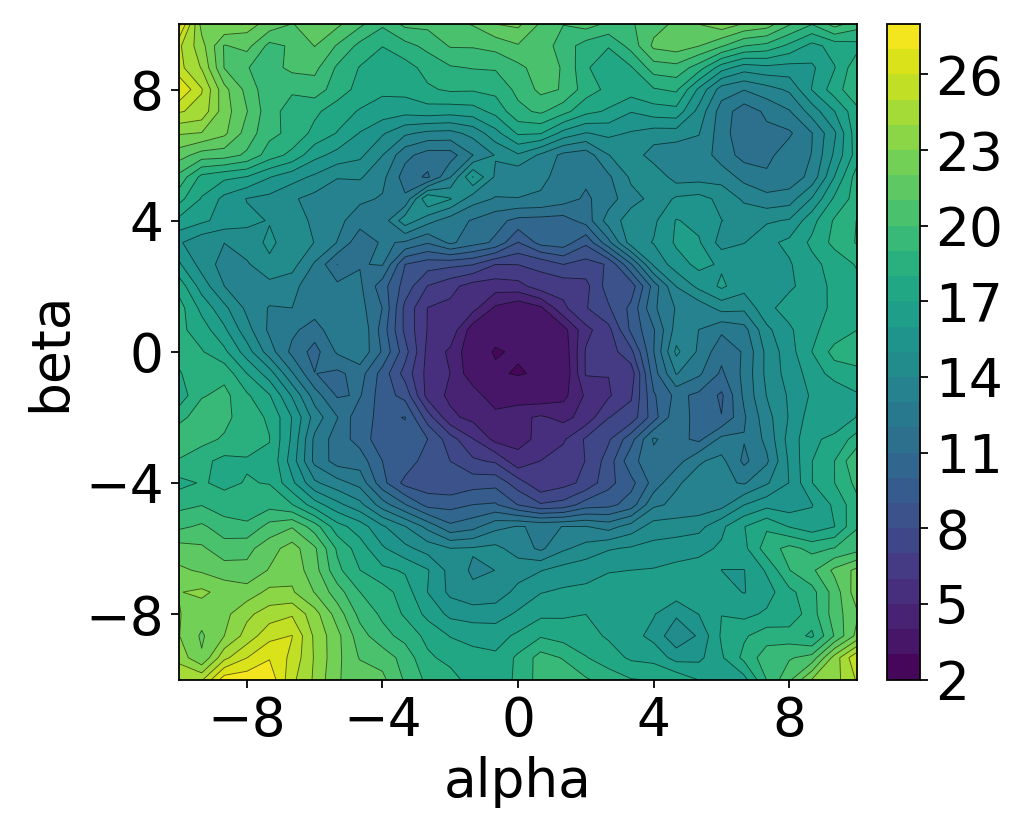} &
\includegraphics[width=0.31\textwidth]{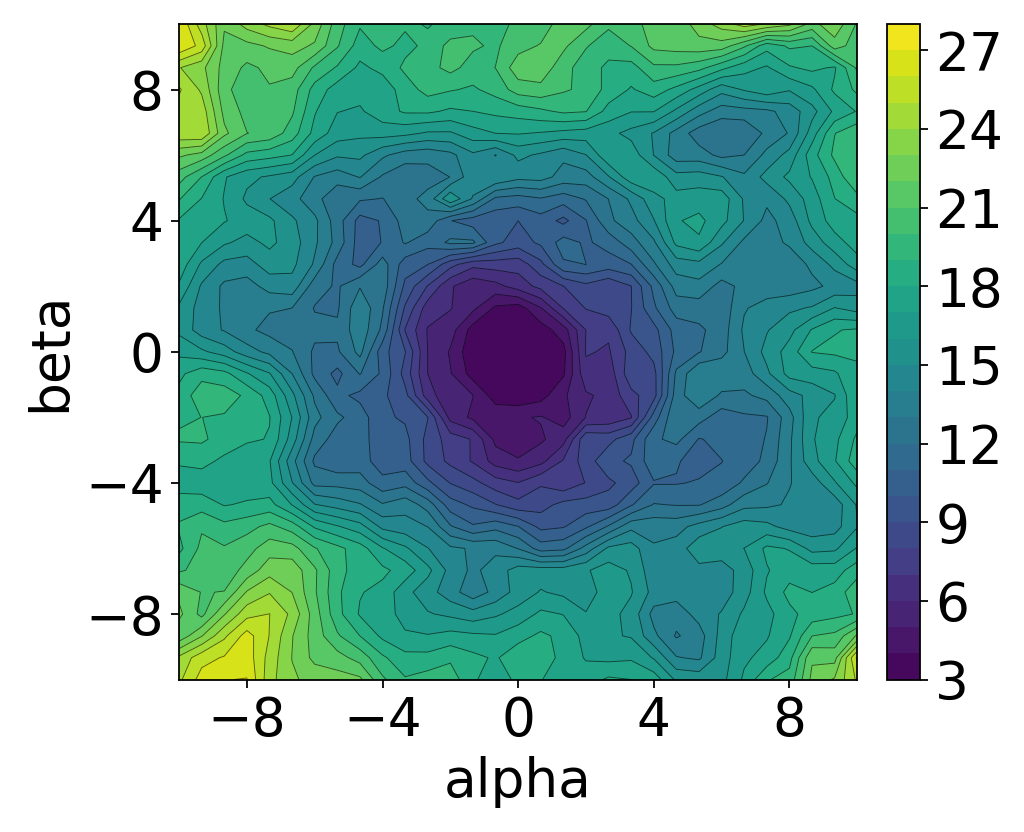} \\
\small Multilingual expert (MMLU-ProX) &

\small PSO (MMLU-ProX) &
\small SAE (MMLU-ProX)

\end{tabular}

\vspace{1mm}
\caption{\textbf{Loss landscapes along shared random directions.}
Each row corresponds to a single task, and each column compares the expert model,
the SAE-merged model, and the PSO-merged model under the same random directions
$(\alpha,\beta)$ in parameter space.}
\label{fig:loss_landscape_sae_vs_pso}
\end{figure*}

Figure~\ref{fig:loss_landscape_sae_vs_pso} visualizes the loss landscapes
of the expert models, the SAE-merged model, and the PSO-merged model
along shared random directions.

On GSM8K (top row), all three models exhibit a low-loss basin centered
around the origin, indicating local stability of the solutions.
Compared to PSO, the SAE-merged model forms a more symmetric and smoothly
varying basin, while the PSO landscape closely resembles that of the math expert.

A similar pattern is observed on MMLU-ProX (bottom row).
The multilingual expert shows a more anisotropic loss surface,
which is largely retained by PSO after merging.
In contrast, SAE produces a more regular and isotropic basin,
suggesting that sparsity-aware optimization reshapes the local loss geometry
rather than inheriting expert-specific structures.

These geometric differences complement the convexity analysis in the main text
and provide an intuitive explanation for SAE’s more consistent performance
improvements over PSO.



\end{document}